\documentclass{article}

\PassOptionsToPackage{numbers, compress}{natbib}

\usepackage[preprint]{neurips_2025}

\usepackage[utf8]{inputenc} 
\usepackage[T1]{fontenc}    
\usepackage{hyperref}       
\usepackage{url}            
\usepackage{booktabs}       
\usepackage{amsfonts}       
\usepackage{graphicx}       
\usepackage{nicefrac}       
\usepackage{multirow}
\usepackage[table]{xcolor}
\usepackage{microtype}      
\usepackage{xcolor}         
\usepackage{xpatch}
\usepackage{amsmath}
\usepackage{xpatch}
\usepackage{subcaption}
\usepackage{wrapfig}
\usepackage{amssymb}
\usepackage{array} 
\makeatletter
\xapptocmd{\NAT@bibsetnum}{\setlength{\leftmargin}{0pt}\setlength{\itemindent}{\labelwidth}\addtolength{\itemindent}{\labelsep}}{}{}
\makeatother

\title{Agentic Feature Augmentation: Unifying Selection and Generation with Teaming, Planning, and Memories}

\author{%
\textbf{Nanxu Gong}\textsuperscript{1}\thanks{Equal contribution}, 
\textbf{Sixun Dong}\textsuperscript{1}\footnotemark[1], 
\textbf{Haoyue Bai}\textsuperscript{1}, 
\textbf{Xinyuan Wang}\textsuperscript{1}, \\
\textbf{Wangyang Ying}\textsuperscript{1}, 
\textbf{Yanjie Fu}\textsuperscript{1} \\
\textsuperscript{1}Arizona State University\\
\texttt{nanxugong@outlook.com}, \\
\texttt{\{sixundong, haoyuebai, xwang735, wying4, yanjie.fu\}@asu.edu}, 
}

\begin{document}

\maketitle

\begin{abstract}

As a widely-used and practical tool, feature engineering transforms raw data into discriminative features to advance AI model performance. However, existing methods usually apply feature selection and generation separately, failing to strive a balance between reducing redundancy and adding meaningful dimensions.
To fill this gap, we propose an agentic feature augmentation concept, where the unification of feature generation and selection is modeled as agentic teaming and  planning. Specifically, we develop a Multi-Agent System with Long and Short-Term Memory (MAGS), comprising a selector agent to eliminate redundant features, a generator agent to produce informative new dimensions, and a router agent that strategically coordinates their actions. We leverage in-context learning with short-term memory for immediate feedback refinement and long-term memory for globally optimal guidance. Additionally, we employ offline Proximal Policy Optimization (PPO) reinforcement fine-tuning to train the router agent for effective decision-making to navigate a vast discrete feature space. Extensive experiments demonstrate that this unified agentic framework consistently achieves superior task performance by intelligently orchestrating feature selection and generation.

\end{abstract}

\section{Introduction}

Feature engineering is effective and practical because it converts complex, noisy raw data into insightful, predictive features to boost AI models' accuracy and efficiency.  The article of ``The Secret Sauce to Winning'' highlighted that when it comes to Kaggle competitions, the most powerful models often come down to feature engineering\footnote{\url{https://www.kaggle.com/discussions/getting-started/540243}}.

The two essential techniques in feature engineering are feature selection (e.g., filtering, wrapper methods)  and feature generation (e.g., transformations, latent embeddings). 
Although feature generation introduces new dimensions, it brings in redundancy. 
Similarly, although feature selection reduces redundancy,  it inadvertently eliminates features that are weak in isolation yet contribute meaningful information through nonlinear interactions. 
Moreover, existing practices often apply selection and generation separately, which can lead to suboptimal models by missing synergistic interactions between them.
In many domain-specific applications (e.g., predictive equipment maintenance), we not just select key sensor signals (e.g.,  vibration frequency, temperature readings, pressure levels collected from industrial machinery), but also generate composite health indicators (e.g., machine health index, failure probability score) from these raw signals, to build better feature representations for accurate prediction.  
This example highlights the need for unifying feature selection and generation.

Existing systems partially address this problem: 
1) feature selection, including filter, wrapper, and embedded methods, are effective in reducing dimensionality~\cite{wang2025towards,ying2025survey,guyon2002gene}. 
Recent advances incorporate reinforcement learning (RL) and generative models to improve automation, fast search, self-identification of selection size~\cite{gong2025neuro,liu2021automated}. 
However, feature selection only filters existing features, cannot create meaningful dimensions, and, thus, risks the loss of hidden interactions and complex patterns needed for more predictive models.
2) feature generation, from manual engineering to AutoML and RL-guided approaches, aims to add meaningful dimensions~\cite{kanter2015deep,tran2016discrete3,wang2022GRFG,wang2023reinforcement,gong2024evolutionary,gong2025unsupervised}. 
However, the combinations or transformations in feature generation can introduce redundant or suboptimal dimensions, thus, failing to retain the most  concise, yet expressive dimensions.
It is appealing to unify feature selection and generation to navigate large feature space and create an expressive, compact, and task-adaptive data representation.

\textbf{Our Insights: an agentic teaming and planning perspective of feature augmentation.} 
Although unification (i.e., unify feature generation and selection) and planning (i.e., plan the path of feature space transformations toward optimal representation in a large discrete space) are challenging in feature engineering, the agent teaming and reasoning abilities of emerging LLM and agentic AI provide a unique perspective for jointly solving unification and planning. 
In particular, we formulate a generic agentic feature augmentation problem.
We show that a feature set can be described as a token sequence; thereafter, feature selection and generation, which transform a feature set into a new one, are framed as generating token sequences, achieved by LLM agents. 
We highlight that the unification of feature selection and generation can be regarded as the teaming of a selector agent and a generator agent. 
We demonstrate that searching for an optimal feature set in a large discrete space can be seen as a task of learning a router agent to direct the trajectory of teaming feature selector and generator agents toward maximizing downstream task performance. 
For efficient transformation, we leverage in-context learning (commercial LLM APIs) along with short-term and long-term memories to learn feature selector and generator agents. 
For global planning, we develop offline PPO reinforcement fine-tuning to learn the router agent for better dataset-specific policies on deciding a binary choice of feature selection or generation. 
Integrating the insights of teaming, routing, reinforcement fine-tuning, and long and short-term memories can develop a learning structure for the router, selector, and generator agents to perform agentic feature augmentation. 

\textbf{Summary of Proposed Approach.} 
Inspired by the above insights, we propose a \textbf{M}ulti-\textbf{A}gent System with Long and Short-Memory to Unify Feature \textbf{G}eneration and \textbf{S}election (\textbf{MAGS}). 
The framework has two goals: 
1) structuring agents; 
2) learning agents. 
To achieve Goal 1, we develop three collaborative agents: a router, a generator, and a selector. The generator is to generate new features to add; the selector is to identify redundant features to eliminate; the router is to decide whether and how to trigger the feature generator or selector based on the state of the current environment. The three agents collaborate iteratively over a path to optimize a feature set. 
To achieve Goal 2, we develop an offline PPO reinforcement fine-tuning method to train the policies of the router agent, so that the router agent is equipped with task-specific reasoning and planning knowledge for more accurate decisions. 
Besides, we develop long and short-term memory mechanisms to facilitate the in-context learning of the generator and selector agents. 
The short-term memory describes the actions in the trajectory of one agentic exploration iteration to allow both selector and generator to refine their actions based on immediate feedback. 
The long-term memory includes high-quality augmented feature sets stored in the history, in order to encourage agents to explore globally-optimal directions. 
Both memories facilitate more stable, informed, and performance-driven feature engineering.

\textbf{Our Contributions: } 
1) We study the problem of feature augmentation and propose an agentic teaming and planning perspective.  
2) We develop a router-selector-generator agentic teaming structure to unify feature generation and selection. 
3) We devise long and short-term memory augmented in-context learning for learning the selector and generator agents, and an offline PPO reinforcement fine-tuning approach for learning the router agent. 
4) Extensive experiments show that such an agentic structure can better identify decision pathways for optimal feature set search in a large discrete space.

\section{Background and Problem Statement}

We present the concepts of agentic feature augmentation to support the design of agents that team feature generation and selection while navigating the complex landscape of feature  space.

\textbf{Operator Set.}
To enable feature crossing and generation, we define the operator set $\mathcal{O} = \{O_1, O_2, \cdots, O_m\}$, including both unary (e.g., square, log, exp) and binary (e.g., add, multiply, subtract) mathematical operations, so that agents can apply these operators to cross original features and generate new features.

\textbf{Feature Set as Token Sequence.}
Given an original feature set $\mathcal{F} = [f_1, f_2, \dots, f_t]$, after applying feature selection and generations, we can obtain an augmented feature set. This augmented feature set can be denoted by a token sequence, e.g., $( f_1 * f_2), (sin f_3), (f_4 - f_5)$. 
To help machines better comprehend feature sets tokenization,  we utilize the postfix expression to change the feature sequence (e.g., $f_1 f_2 *, f_3 sin, f_4 f_5 -$).

\textbf{Feature Generation.}
Feature generation aims to generate new features by applying mathematical operations. Given a raw feature set [$a, b$], a new feature set [$a+b, a/b, log b$] can be generated by crossing features with operators. 

\textbf{Feature Selection.}
Feature selection aims to identify the most informative features while removing redundant or irrelevant ones. 
For example, given a feature set $[a, b, c, d]$, a selection operation may retain only $[a, b]$ based on their contribution to the downstream task. 

\begin{figure*}[t]
    \centering
    \includegraphics[width=0.98\linewidth]{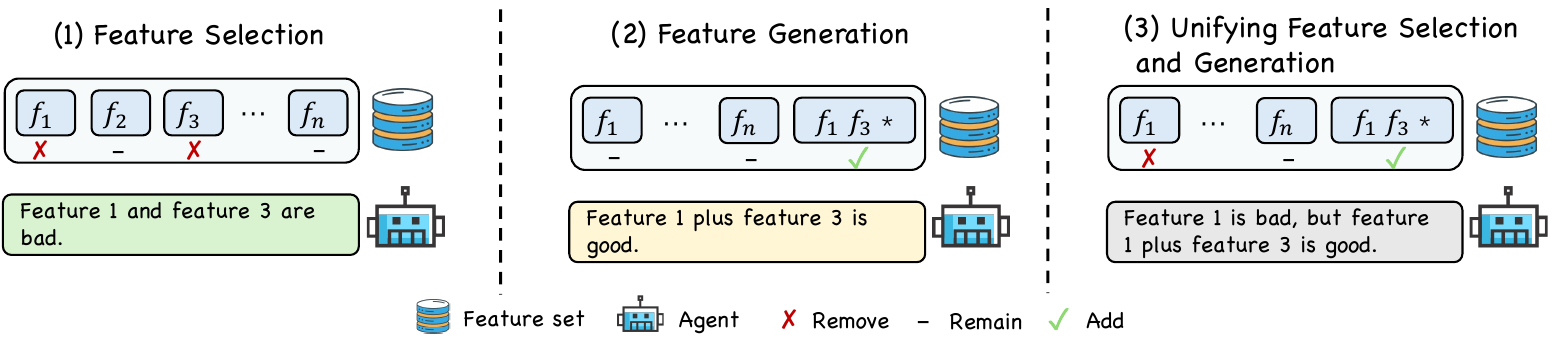}
    \caption{Example of feature selection, feature generation, and unifying feature selection and generation.}
    \label{fig:problem}
    \vspace{-0.2cm}
\end{figure*}

Unlike prior work that treats feature transformation and selection as separate stages, our objective is to unify both operations within an iterative agentic process to augment the feature representation of data (\textbf{Figure \ref{fig:problem}}). 

\textbf{Short-Term Memory} 
is the actions in the trajectory of one agentic exploration iteration without restart, for instance, a sequence of records on selection or generation, how to select or generate,  augmented feature token sequences, and corresponding performance. 

\textbf{Long-Term Memory}  is defined as demonstrations of high-quality augmented feature sets in historical memory. Both long-term and short-term memories serve as analogy learning demonstrations to guide agents to self-optimize feature augmentation policies. 

\noindent\textbf{The Agentic Feature Augmentation Problem.}
Given an original feature set $\mathcal{F}_0$, we aim to develop collaborative agents to coordinate feature selection and generation to augment data representation.  Formally, considering the existence of the optimization process structured by $m$ total iterations, and each iteration consists of $n$ agentic decision steps.
In particular, in the $i$-th iteration, the input feature set is denoted by $\mathcal{F}_{i,0}$. In the $j$-th step of the $i$-th  iteration, a router agent determines whether to conduct feature generation or selection . 
Based on this binary decision, a feature transformer agent or a feature selector agent produces an updated feature set $\mathcal{F}_{i,j}$.
The goal is to identify the optimal feature set that maximizes downstream task performance through iterations and steps given by:
\begin{equation}
    \mathcal{F}^* = \arg\max_{i,j \in I,J} \mathcal{S}(\mathcal{F}_{i,j}, \mathcal{Y}),
\end{equation}
where $\mathcal{F}_{i,j}$ denotes the feature set generated at step $j$ of iteration $i$, $\mathcal{Y}$ is the ground-truth label set, and $\mathcal{S}(\cdot)$ is a task-specific scoring function that evaluates the quality of a given feature set.
\section{Agentic Feature Augmentation}

\subsection{Framework Overview: The Router, Feature Selector, Feature Generator Agents}
\begin{figure*}[t]
    \centering
    \includegraphics[width=0.98\linewidth]{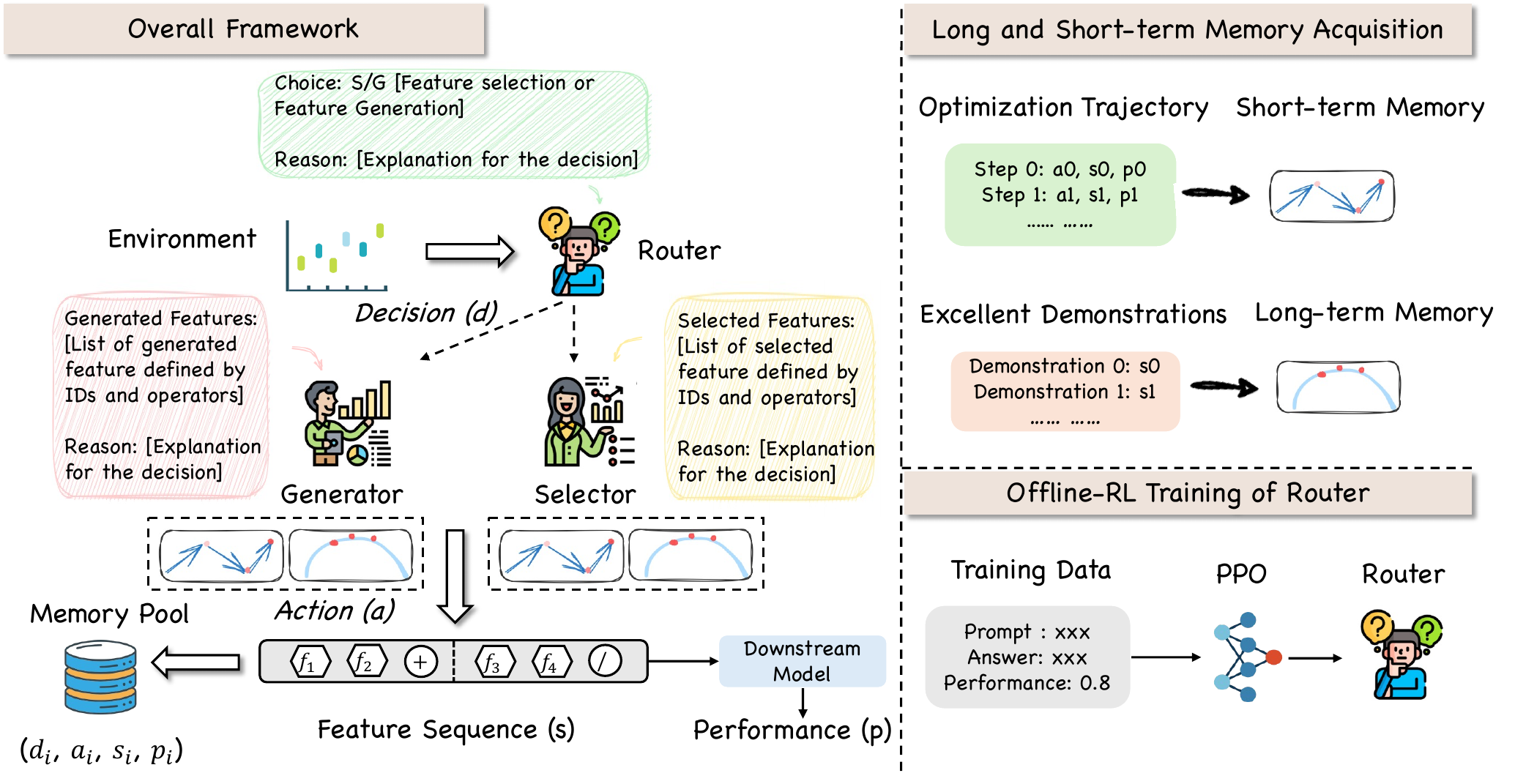}
    \caption{Framework overview. The left section illustrates the overall framework, where a router is first employed to determine whether to perform feature generation or selection , followed by the execution of the corresponding operation. The right section depicts the construction of long and short-term memory representations, along with the offline reinforcement learning process used to train the router.}
    \label{fig:framework}
    \vspace{-0.2cm}
\end{figure*}

Figure~\ref{fig:framework} shows the three technical components of our method: (1) an agentic feature augmentation framework to unify feature generation and selection,  (2) a dual memory mechanism for acquiring long and short-term experiences, and  (3) an offline reinforcement learning module for fine-tuning the router for choosing selection or generation agents. 

Firstly, we develop a multi-agent iterative collaboration framework, where a router agent is to analyze input data to decide whether to reduce or add features,  a feature generator agent is to cross features to generate new features,  and a feature selector agent is to remove redundant features. 
This framework iteratively transforms feature sets.
In this process, a feature set is tokenized by a postfix expression named feature token sequence, and is associated with the performance of a downstream task. 
We store the binary decision (generation or selection), generation or selection actions (how to generate or select), augmented feature token sequences, and their associated performances.
Secondly, it is critical to expose agents to informative experiences to learn better. We introduce two concepts: long-term memory and short-term memory by categorizing historical agentic feature augmentation experiences.  Short-term memory is to advance local optimization perceptions of selector/generator agents while long-term memory is to improve agentic global optimization perception when pursuing feature augmentation. 
Thirdly, the router agent, iteratively and dynamically deciding whether to add or reduce features, steers the optimization pathway towards a more global optima in automated feature augmentation. 
As a high-level controller that coordinates feature generation and selection, the router is further fine-tuned via offline reinforcement learning to align with the global optimization objective. This enables the system to maintain global reward alignment while preserving local decision effectiveness.




\subsection{Agent Structure and Collaboration Design for Agentic Feature Augmentation }
Our idea is to view agentic feature augmentation from the perspective of unifying feature generation and selection as an adaptive and collaborative learning machine.  To this end, we design an LLM agent-based collaborative tasking structure, including a router agent, a generator agent, and a selector agent. The router agent is to decide the next action: reducing features via selection or adding features via generation.  The generator agent is to conduct feature crossing to generate meaningful dimensions. The selector agent is to conduct feature selections to eliminate redundant dimensions. 

\noindent\textbf{The Router Agent.} The router agent is an iterative path planner agent that adaptively decides whether to perform feature generation or feature selection at each action step. 
In particular, given the current feature set $\mathcal{F}_{i,j}$, the router agent analyzes the feature sets and state, then chooses generation or selection based on choice policies. 

\noindent\textbf{The Generator Agent.}  
When the router agent chooses feature generation, the generator agent is activated to analyze the current feature set, select which features to cross, decide which operator in the predefined operator set $\mathcal{O}$  to cross the selected features,  and finally generate new features.  The augmented feature set is updated by:
\[
\mathcal{F}_{i,j+1} = \mathcal{F}_{i,j} \cup O_T(\mathcal{F}_{i,j}).
\]

\noindent\textbf{The Selector Agent.}  
When the router agent chooses feature selection, the selector agent is activated to evaluate the relevance of features in $\mathcal{F}_{i,j}$. It generates a binary mask $\mathbf{s}_{i,j}$ that filters out redundant features:
\[
\mathcal{F}_{i,j+1} = \mathcal{F}_{i,j} \odot \mathbf{s}_{i,j},
\]
where $\odot$ denotes column-wise masking. The selector balances informativeness and compactness, and avoids oversized features during iterative augmentations.

\noindent\textbf{Memory Pool.}  
To leverage past experiences for better in-context learning, we develop a memory pool that stores all the exploration iterations and corresponding action sequences in each iteration. The agents explore multiple iterations. Each iteration includes a sequence of actions. Each action consists of four components: 
(1) the binary decision made by the router (i.e., feature selection or feature generation), 
(2) the steps taken by the generator or selector (i.e., how to select or generate), (3) the resulting augmented feature set $\mathcal{F}_{i,j}$, represented by a postfix token sequence (e.g., $f1 f2 -, f3 sin, f4 f5 *$), and 
(4) the downstream task performance evaluated by the scoring function $\mathcal{S}(\cdot)$.

\textbf{Agentic Collaboration.} 
Throughout the feature augmentation process, our multi-agent system operates in an iteration with restart fashion to progressively improve the feature representation space. 
In particular, we iterate multiple iterations. 
Each iteration starts by regarding the given original feature set as an initialized feature set.  
Each iteration is associated with multiple actions. 
In each action, the router agent firstly makes a binary decision on feature generation or selection,  based on the state of the current feature set and task contexts. 
Later, the generator or selector agent is triggered to pick a generation (e.g., $f_1 * f_2$) or selection (e.g., remove $f_1$) action to the current feature set.
This process is seen as a nested loop with multiple iterations, where each iteration has multiple actions. so that the feature space is adaptively expanded, pruned, and reorganized. 
Through this collaborative process, the agents explore a diverse range of feature compositions while remaining aligned with the objective of optimizing downstream tasks. 

\subsection{Learning Generator, Selector, and Routing Agents}
We next discuss how to learn the policies of the generator, selector, and routing agents. 

\subsubsection{Leveraging Long and Short-Term Memory for In-context Learning of The Selector and Generator Agents}
Next, to further support long-term planning, promote knowledge reuse, and stabilize the optimization dynamics, we propose two concepts: \textit{short-term memory} and \textit{long-term memory}.
We leverage the two memories to enhance the short-term adaptability and long-term generalization of each agent for better decision quality.

\noindent\textbf{Short-Term Memory.}  
The short-term memory is defined as the agent-specific action sequence within the current iteration.
The records of each action include: 1) the specific agent's operations on how to select or generate; 2) the resulting feature sequences, and 3) their associated performance scores. 
Such local memory allows the agent to adaptively and iteratively update its policies based on local and recent feedback, enabling adaptive behavior and fine-grained correction during the ongoing feature augmentation optimization loop.

\noindent\textbf{Long-Term Memory.}  
The long-term memory is defined as a set of high-performance samples in the memory pool based on downstream task accuracies. 
To ensure long-term memory to be high-quality and diverse, we introduce randomness to randomly sample K demonstrations from the top-performing experiences, instead of simply selecting fixed top-$K$ demonstrations. 
The random sampling encourages agents to learn the analogy of successful patterns while avoiding overfitting to a narrow set of optimal cases. 
The demonstrations in the long-term memory are utilized in agentic prompts as contextual knowledge, in order to help agents learn effective feature generation or selection policies.

\textbf{Finally}, the integration of short-term and long-term memory introduces computational benefits: 1) short-term memory enables each agent to adapt its actions in real time by leveraging immediate feedback within an iteration, promoting fine-grained optimization and stability; 
2) long-term memory, on the other hand, provides strategic guidance by distilling high-quality historical knowledge into reusable prompts, facilitating efficient exploration and preventing overfitting to suboptimal patterns.  Together, the two complementary memories balance exploitation and exploration, improve decision consistency across iterations, and ultimately boost the overall quality of the generated feature sets.

\subsubsection{Offline Reinforcement Finetuning of The Router Agent for Optimal Planning}
The router plays a critical role because it steers the directed path of feature augmentation toward the optimal feature space. To learn a better router agent to make accurate task-specialized decisions, we frame the router agent as a policy network that selects an action $a_t \in \{\texttt{generation}, \texttt{selection}\}$ based on the current feature state and contextual information, and develop an offline reinforcement fine-tuning method for better choice modeling. 

\noindent\textbf{Offline Training Data Collection.}  
We collect a training dataset composed of triplets of $(\texttt{prompt}, \texttt{answer}, \texttt{score})$, where 1) \texttt{prompt} encodes the current environment state, such as descriptive statistics (e.g., mean, std, max, min) of the features and textual instructions.
2) \texttt{answer} is the routing decision made by the router (i.e., whether to invoke the generator or selector and the reason for the decision).
3) \texttt{score} represents the downstream task performance resulting from that decision, serving as the reward signal.

\noindent\textbf{Optimization Strategy.}  
We develop a Proximal Policy Optimization (PPO) based  algorithm to fine-tune the router’s policy using the offline data. PPO stabilizes training by preventing large updates to the policy and encouraging monotonic improvement. At each update step, the router receives a prompt as input and generates a routing decision. The policy is then optimized to increase the expected reward (i.e., downstream performance) associated with that decision, while staying close to the behavior policy reflected in the dataset.

Through this process, the router learns to make context-aware decisions that adapt to both dataset characteristics and evolving feature space. This enables more effective choice modeling of feature generation and selection and the optimization path planning toward optimal feature augmentation , ultimately improving the quality of the learned features. 

\section{Experimental Results}

To comprehensively validate the effectiveness of the proposed method, we conduct various experiments to answer the following research questions: \textbf{RQ1:} Is the proposed method effective in improving downstream task performance compared to existing feature generation and selection approaches?
\textbf{RQ2:} Does offline reinforcement learning fine-tuning improve the decision quality of the router and contribute to overall performance?
\textbf{RQ3:} Are the proposed dual memory mechanisms beneficial in guiding the optimization process and enhancing feature quality?
\textbf{RQ4:} Is the method robust when integrated with different downstream models?
\textbf{RQ5:} Can the method produce an interpretable and traceable feature set?

\subsection{Experimental Setup}

We conduct experiments on 6 publicly available datasets, which are categorized into two prediction tasks: classification and regression. For all tasks, we use Random Forest as the default downstream model. Classification performance is evaluated using  F1 score and Accuracy, while regression performance is assessed using 1 - Mean Squared Error (1-MSE) and the coefficient of determination ($R^2$). We compare our proposed method with a diverse set of baselines, which fall into two categories: feature generation and feature selection. The feature generation methods include ERG, AFT \cite{horn2020autofeat}, NFS \cite{chen2019neural}, TTG \cite{khurana2018feature}, GRFG \cite{wang2022GRFG}, and ELLM-FT \cite{gong2024evolutionary}; the feature selection methods include KBEST \cite{yang1997comparative}, LASSO \cite{tibshirani1996regression}, RFE \cite{granitto2006recursive}, SARLFS \cite{liu2021efficient}, MARLFS \cite{liu2021automated}, and FSNS \cite{gong2025neuro}.
For the router, we employ \textit{LLaMA-3.2-3B}, while the selector and generator agents are implemented using \textit{GPT-3.5-Turbo}.
More details about the datasets, baselines, and hyperparameters are provided in Appendix~\ref{ap:ex}.

To comprehensively evaluate the effectiveness of each component, we introduce four ablation variants of MAGS:
1) \textit{w/o RL}: Removes the offline reinforcement learning fine-tuning for the router module, using only the initial policy without learning from performance feedback.
2) \textit{w/o Long}: Disables the long-term memory mechanism, preventing agents from incorporating high-quality demonstrations from past iterations into their prompt context.
3) \textit{w/o Short}: Disables the short-term memory mechanism, so agents no longer adapt their behavior based on recent decision-feedback trajectories within an iteration.
4) \textit{w/o Router}: Removes the router module and replaces its decision with a uniformly random choice between generation and selection at each step. This variant evaluates the importance of adaptive, context-aware scheduling.

\begin{table*}[t]
\centering
\small
\renewcommand{\arraystretch}{1.2}
\resizebox{\linewidth}{!}{%
\begin{tabular}{lccccccccccccc}
\toprule
\multirow{2}{*}{\textbf{Method}} & \multirow{2}{*}{\textbf{Type}} 
& \multicolumn{2}{c}{\textbf{german\_credit}} 
& \multicolumn{2}{c}{\textbf{svmguide3}} 
& \multicolumn{2}{c}{\textbf{messidor}} 
& \multicolumn{2}{c}{\textbf{openml\_586}} 
& \multicolumn{2}{c}{\textbf{openml\_589}} 
& \multicolumn{2}{c}{\textbf{openml\_607}} \\
\cmidrule(lr){3-4} \cmidrule(lr){5-6} \cmidrule(lr){7-8} \cmidrule(lr){9-10} \cmidrule(lr){11-12} \cmidrule(lr){13-14}
&  & F1~$\uparrow$ & Acc~$\uparrow$ & F1~$\uparrow$ & Acc~$\uparrow$ & F1~$\uparrow$ & Acc~$\uparrow$ & 1-MSE~$\uparrow$ & R$^2$~$\uparrow$ & 1-MSE~$\uparrow$ & R$^2$~$\uparrow$ & 1-MSE~$\uparrow$ & R$^2$~$\uparrow$ \\
\midrule
ERG       & G & 0.744 & 0.764 & 0.828 & 0.842 & 0.705 & 0.705 & 0.893 & 0.893 & 0.881 & 0.879 & 0.881 & 0.879 \\
AFT \cite{horn2020autofeat}       & G & 0.746 & 0.763 & 0.830 & 0.844 & 0.673 & 0.673 & 0.895 & 0.894 & 0.883 & 0.880 & 0.881 & 0.878 \\
NFS \cite{chen2019neural}       & G & 0.747 & 0.766 & 0.830 & 0.845 & 0.700 & 0.700 & 0.895 & 0.894 & 0.885 & 0.883 & 0.879 & 0.877 \\
TTG \cite{khurana2018feature}       & G & 0.750 & 0.770 & 0.831 & 0.846 & 0.702 & 0.702 & 0.895 & 0.894 & 0.880 & 0.878 & 0.881 & 0.878 \\
GRFG \cite{wang2022GRFG}      & G & 0.746 & 0.767 & 0.832 & 0.847 & 0.692 & 0.692 & 0.895 & 0.894 & 0.880 & 0.878 & 0.881 & 0.879 \\
ELLM-FT \cite{gong2024evolutionary}   & G & 0.756 & 0.774 & \textbf{0.845} & \textbf{0.855} & 0.724 & 0.724 & 0.926 & 0.925 & 0.909 & 0.907 & 0.890 & 0.888 \\
\midrule
KBEST \cite{yang1997comparative}     & S & 0.681 & 0.715 & 0.765 & 0.791 & 0.589 & 0.593 & 0.814 & 0.804 & 0.758 & 0.745 & 0.762 & 0.771 \\
LASSO \cite{tibshirani1996regression}     & S & 0.676 & 0.705 & 0.756 & 0.783 & 0.620 & 0.624 & 0.844 & 0.836 & 0.820 & 0.810 & 0.797 & 0.805 \\
RFE \cite{granitto2006recursive}       & S & 0.654 & 0.690 & 0.794 & 0.815 & 0.641 & 0.645 & 0.819 & 0.808 & 0.764 & 0.751 & 0.783 & 0.792 \\
SARLFS \cite{liu2021efficient}    & S & 0.678 & 0.705 & 0.809 & 0.827 & 0.677 & 0.680 & 0.794 & 0.787 & 0.718 & 0.703 & 0.750 & 0.761 \\
MARLFS \cite{liu2021automated}    & S & 0.665 & 0.715 & 0.820 & 0.839 & 0.601 & 0.601 & 0.801 & 0.791 & 0.755 & 0.741 & 0.775 & 0.781 \\
FSNS  \cite{gong2025neuro}     & S & 0.739 & 0.760 & 0.823 & 0.843 & 0.682 & 0.684 & 0.842 & 0.833 & 0.834 & 0.824 & 0.789 & 0.798 \\
\rowcolor{gray!15}
\midrule
\textbf{MAGS} & \textbf{B} & \textbf{0.763} & \textbf{0.776} & 0.840 & \textbf{0.855} & \textbf{0.754} & \textbf{0.754} & \textbf{0.955} & \textbf{0.956} & \textbf{0.938} & \textbf{0.937} & \textbf{0.947} & \textbf{0.948} \\
\bottomrule
\end{tabular}
}
\caption{Overall comparison across classification (F1~$\uparrow$, Accuracy~$\uparrow$) and regression (1-MSE~$\uparrow$, $R^2$~$\uparrow$) datasets. Best value in each column is bolded. Type: G (generation), S (selection), B (both).}
\label{tab:overall}
\vspace{-0.4cm}
\end{table*}

\subsection{Overall Comparison (RQ1)}

Table~\ref{tab:overall} reports the performance of MAGS and all baselines across 6 benchmark datasets. Overall, MAGS consistently outperforms all competing methods on most datasets and evaluation metrics, demonstrating the effectiveness of jointly modeling feature selection and generation within a unified framework. This unified design allows the system to simultaneously explore expressive feature compositions and eliminate redundant or noisy features, enabling better trade-offs between representation power and generalization.
An interesting observation is that ELLM-FT, a strong feature generation baseline, performs competitively on certain datasets. However, its performance varies significantly, especially on noisy or high-dimensional datasets such as openml\_607, where uncontrolled feature generation may introduce irrelevant features and degrade performance. In contrast, MAGS exhibits stronger robustness by adaptively selecting relevant features while still benefiting from expressive generation.
These results highlight the value of treating feature engineering as a coordinated decision-making process, rather than as a sequential or isolated pipeline. By unifying selection and generation, MAGS is better positioned to construct high-quality, task-specific feature spaces that are both interpretable and model-friendly.

\subsection{Ablation Study (RQ2 and RQ3)}

\begin{wraptable}{r}{0.55\textwidth}
\vspace{-4.5mm}
\centering
\small
\renewcommand{\arraystretch}{1.2}
\setlength{\tabcolsep}{4pt}
\begin{tabular}{lcccc}
\toprule
\textbf{Dataset} & \textbf{S (w/o RL)} & \textbf{S} & \textbf{G (w/o RL)} & \textbf{G} \\
\midrule
svmguide3   & \textbf{62\%} & 36\%$\downarrow$ & 38\% & \textbf{64\%}$\uparrow$ \\
messidor    & 40\% & 35\%$\downarrow$ & \textbf{60\%} & \textbf{65\%}$\uparrow$ \\
openml\_589 & \textbf{66\%} & \textbf{76\%}$\uparrow$ & 34\% & 24\%$\downarrow$ \\
openml\_607 & 43\% & \textbf{53\%}$\uparrow$ & \textbf{57\%} & 47\%$\downarrow$ \\
\bottomrule
\end{tabular}
\caption{Selection and generation decisions before and after offline RL fine-tuning.  arrows indicate increase ($\uparrow$) or decrease ($\downarrow$). The dominant is highlighted in bold.}
\label{tab:selection_generation}
\vspace{-0.4cm}
\end{wraptable}

\begin{figure}[h]
    \centering
    \begin{minipage}[b]{0.24\textwidth}
        \centering
        \includegraphics[width=\textwidth]{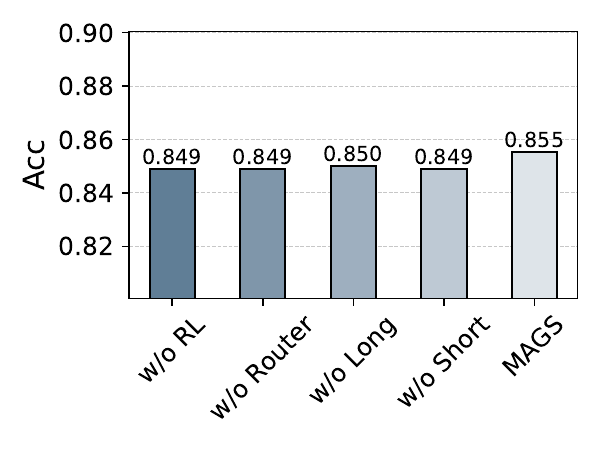}
        \caption*{(a) svmguide3}
    \end{minipage}
    \begin{minipage}[b]{0.24\textwidth}
        \centering
        \includegraphics[width=\textwidth]{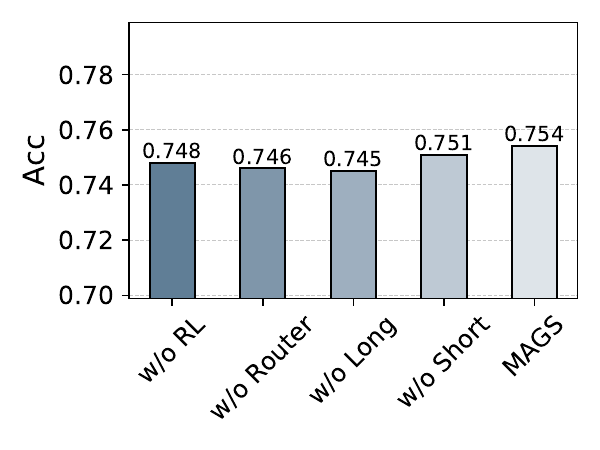}
        \caption*{(b) messidor}
    \end{minipage}
    \begin{minipage}[b]{0.24\textwidth}
        \centering
        \includegraphics[width=\textwidth]{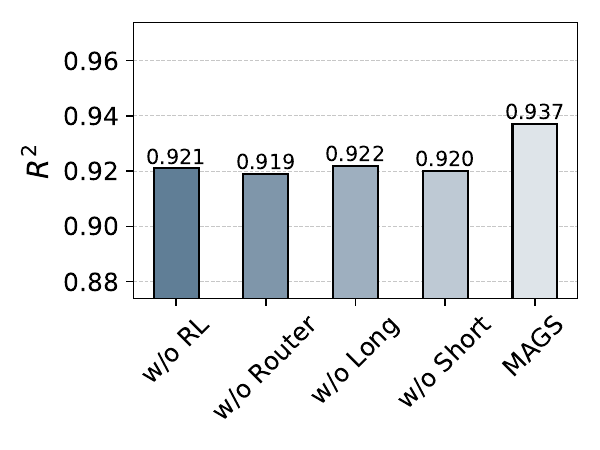}
        \caption*{(c) openml\_589}
    \end{minipage}
    \begin{minipage}[b]{0.24\textwidth}
        \centering
        \includegraphics[width=\textwidth]{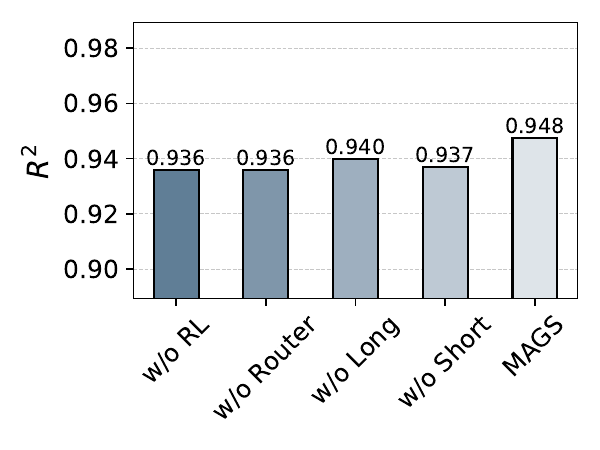}
        \caption*{(d) openml\_607}
    \end{minipage}
    
    \caption{Ablation study. We compare the performance of the variant models with MAGS on 4 datasets.}
    \vspace{-0.25cm}
    \label{fig:ab}
\end{figure}

To assess the contribution of each component in our proposed framework, we conduct ablation studies with four variant models: \textit{w/o RL}, \textit{w/o Router}, \textit{w/o Long}, and \textit{w/o Short}. These variants are designed to evaluate the roles of offline reinforcement learning fine-tuning, the routing mechanism, long-term memory, and short-term memory, respectively. Figure~\ref{fig:ab} presents the results across four representative datasets, with classification tasks evaluated using accuracy and regression tasks using $R^2$ scores. From the results, we draw two key insights:
\textit{1) Dual memory mechanisms significantly enhance feature optimization.}  
Both long-term and short-term memory provide complementary benefits: long-term memory injects high-quality, diverse examples into the prompt to guide exploration toward globally promising directions, while short-term memory allows the agent to react to recent feedback and refine its strategy locally. Removing either memory module leads to consistent performance degradation, highlighting their synergistic effect in stabilizing and guiding the optimization process.
\textit{2) Fine-tuned router is crucial for effective decision scheduling.}  
The router agent, when fine-tuned via offline reinforcement learning, learns to adaptively determine whether to perform generation or selection based on context. Removing RL fine-tuning (\textit{w/o RL}) or replacing the router with random routing decisions (\textit{w/o Router}) results in a substantial performance drop. This demonstrates that task-aware, context-sensitive scheduling is essential for navigating the feature engineering space effectively, and validates the use of reinforcement learning for learning a high-quality routing policy.

To further analyze the impact of RL fine-tuning on the router’s behavior, we conduct an additional experiment that examines the distribution of routing decisions across datasets. Table~\ref{tab:selection_generation} reports the percentage of steps where the router selected generation versus selection operations, both with and without RL fine-tuning.
We observe that RL fine-tuning enables the router to develop task-specific scheduling preferences. For instance, on low-dimensional datasets, the fine-tuned router tends to favor generation more frequently, aiming to enrich the feature space with expressive combinations. In contrast, on noisy or high-dimensional datasets, such as \textit{openml\_589}, the router prioritizes selection, focusing on eliminating irrelevant or redundant features to improve generalization.
These results indicate that reinforcement learning allows the router to go beyond naive or uniform scheduling and develop an adaptive strategy that aligns with dataset characteristics, thereby contributing to the overall effectiveness of the feature engineering process.

\subsection{Robustness Check (RQ4)}

\begin{wraptable}{r}{0.55\textwidth}
\vspace{-0.4cm}
\centering
\small
\renewcommand{\arraystretch}{1.2}
\setlength{\tabcolsep}{4pt}
\begin{tabular}{lcccccc}
\toprule
\textbf{Dataset} & \textbf{GB} & \textbf{KNN} & \textbf{LR} & \textbf{MLP} & \textbf{SVM} & \textbf{RF} \\
\midrule
TTG     & 0.699 & 0.642 & 0.713 & 0.733 & 0.712 & 0.702 \\
GRFG    & 0.699 & 0.646 & 0.746 & 0.729 & 0.750 & 0.692 \\
ELLM-FT & 0.705 & 0.699 & 0.752 & 0.743 & 0.752 & 0.724 \\
SARLFS  & 0.624 & 0.645 & 0.676 & 0.701 & 0.710 & 0.680 \\
MARLFS  & 0.628 & 0.637 & 0.714 & 0.680 & 0.662 & 0.601 \\
FSNS    & 0.671 & 0.671 & 0.723 & 0.755 & 0.714 & 0.684 \\
\rowcolor{gray!15}
\midrule
\textbf{MAGS}    & \textbf{0.746} & \textbf{0.753} & \textbf{0.759} & \textbf{0.756} & \textbf{0.761} & \textbf{0.754} \\
\bottomrule
\end{tabular}
\caption{Robustness check. We compare the performance evaluated by different downstream ML models on the dataset \textit{messidor}.}
\label{tab:rc}
\vspace{-0.4cm}
\end{wraptable}

To evaluate the robustness of MAGS across different downstream tasks, we test it on the \textit{messidor} dataset using 6 machine learning models: Gradient Boosting (GB), k-Nearest Neighbors (KNN), Logistic Regression (LR), Multi-Layer Perceptron (MLP), Support Vector Machine (SVM), and Random Forest (RF). Table~\ref{tab:rc} reports the classification accuracy for each model.
Across all classifiers, MAGS consistently achieves superior performance compared to all baseline methods, demonstrating its strong adaptability and robustness in diverse modeling contexts. This consistent improvement suggests that MAGS is capable of generating feature sets that are not only broadly applicable but also effectively tailored to different model-specific characteristics and learning dynamics. Particularly noteworthy is its performance on the KNN classifier, where MAGS surpasses the best-performing baseline by approximately 5.4\% in accuracy. This significant margin further underscores the ability of MAGS to enhance performance even in challenging scenarios where traditional methods often struggle. Such empirical evidence supports the practical utility of MAGS as a versatile feature generation framework that generalizes well across tasks and models.
\begin{wrapfigure}{r}{0.5\textwidth}
    \centering
    \includegraphics[width=0.98\linewidth]{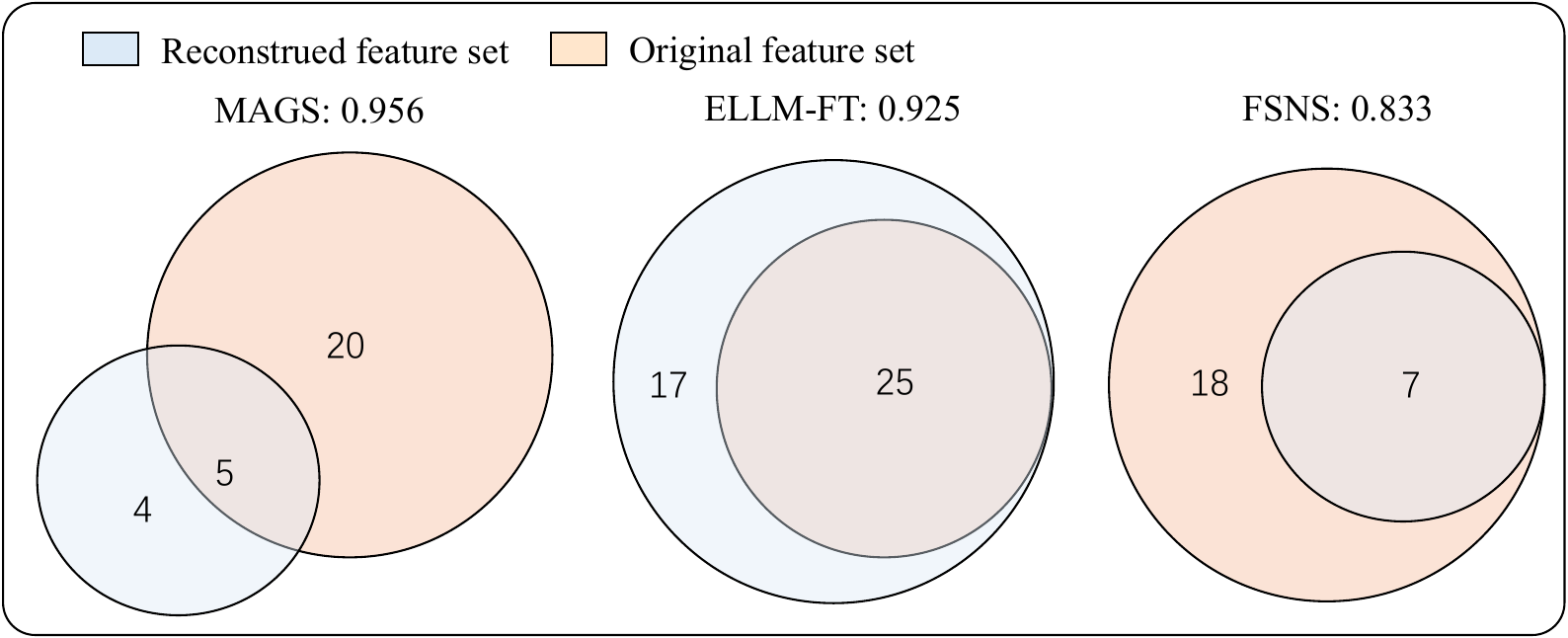}
    \caption{Case study. We visualize the reconstructed feature set on the dataset openml\_586.}
    \label{fig:case}
    \vspace{-0.2cm}
\end{wrapfigure}

\subsection{Case Study: Traceability and Explainability (RQ5)}

To evaluate the traceability and explainability of the feature space, we conduct a case study on the \textit{openml\_586} dataset by visualizing the reconstructed feature sets produced by ELLM-FT, FSNS, and MAGS, as shown in Figure~\ref{fig:case}.
We observe that MAGS generates 4 new features while removing 20 redundant ones. In comparison, ELLM-FT produces 17 new features with no feature selection, and FSNS removes 18 features without performing any generation. This illustrates that MAGS achieves a better balance between feature expansion and compression, selectively introducing expressive combinations while effectively pruning irrelevant or redundant inputs.




\section{Related Work}

\noindent{\bf Feature Generation and Selection.}
Feature generation and selection are fundamental techniques in tabular data-centric AI. Feature selection focuses on identifying the most relevant attributes while removing redundant ones. Traditional methods are typically categorized into filter, wrapper, embedded, and hybrid methods \cite{wang2025towards,ying2025survey}. Filter-based methods are efficient but often overlook feature dependencies, while wrapper and embedded methods consider interactions at the cost of increased computation \cite{guyon2002gene}. Recent advances include generative feature selection \cite{gong2025neuro,ying2024feature,ying2024revolutionizing}, and reinforcement learning (RL)-based frameworks that formulate selection as a sequential decision-making process to enable dynamic adaptation and optimization \cite{liu2021automated,wang2024knockoff}.
Feature generation constructs new features through mathematical operations to enhance data representation. Traditional methods include manual engineering and domain-specific feature design \cite{wang2025towards}. Automated techniques, such as expansion-reduction and evolutionary-evaluation, aim to discover effective features while reducing human effort \cite{kanter2015deep,tran2016discrete3}. More recent work explores reinforcement learning to simulate human-like trial-and-error strategies and optimize generation sequences \cite{wang2022GRFG,ying2023self,ying2024topology,hu2024reinforcement} and generative AI to encode feature knowledge into latent spaces for efficient reuse and exploration \cite{wang2023reinforcement,ying2024unsupervised}.


\noindent\textbf{Multi-Agent Systems}
Multi-agent systems (MAS) have served as a paradigm for modeling distributed intelligence. Recent work has extended traditional MAS into LLM-based multi-agent systems (LLM-MAS), enabling more flexible, communicative, and knowledge-rich agents. Compared to single-agent systems, LLM-MAS benefit from role specialization, inter-agent communication, and collective decision-making~\cite{wang2025mixllm}, allowing them to more effectively tackle tasks such as software development \cite{hong2023metagpt}, multi-robot planning \cite{mandi2024roco}, science debates \cite{xiong2023examining}, and policy simulations \cite{xiao2023simulating}. Researchers have proposed structured frameworks like MetaGPT \cite{hong2023metagpt} and CAMEL \cite{li2023camel}, which encode human workflows or employ inception prompting to guide agent behavior. Studies further analyze communication paradigms (cooperative, debate, competitive), agent profiling strategies (pre-defined, model-generated), and capability acquisition mechanisms including memory, self-evolution, and dynamic generation \cite{guo2024large,wang2024macrec}. Our work builds on this foundation, leveraging multi-agent coordination to unify feature generation and selection.

\section{Conclusion Remarks}

In this work, we present an agentic feature augmentation framework  that integrates feature generation and feature selection within a multi-agent system. By leveraging the collaboration among a router, a generator, and a selector, our method adaptively schedules and applies generation or selection operations in a task-aware manner. The router is fine-tuned via offline reinforcement learning to make globally optimal decisions, and dual memory mechanisms are introduced to guide agent behaviors at both short-term and long-term timescales. Through extensive experiments on diverse benchmark datasets, we demonstrate that our framework consistently outperforms existing methods in terms of predictive performance, robustness across models, and interpretability of the constructed feature space. Our results highlight the effectiveness of unifying generation and selection under a collaborative, memory-enhanced decision-making process.

\bibliographystyle{plainnat}  
\small
\bibliography{refer}
\normalsize
\newpage
\appendix

\section{Prompt Templates for Agent Specialization}

\begin{figure*}[h]
    \centering
    \includegraphics[width=0.8\linewidth]{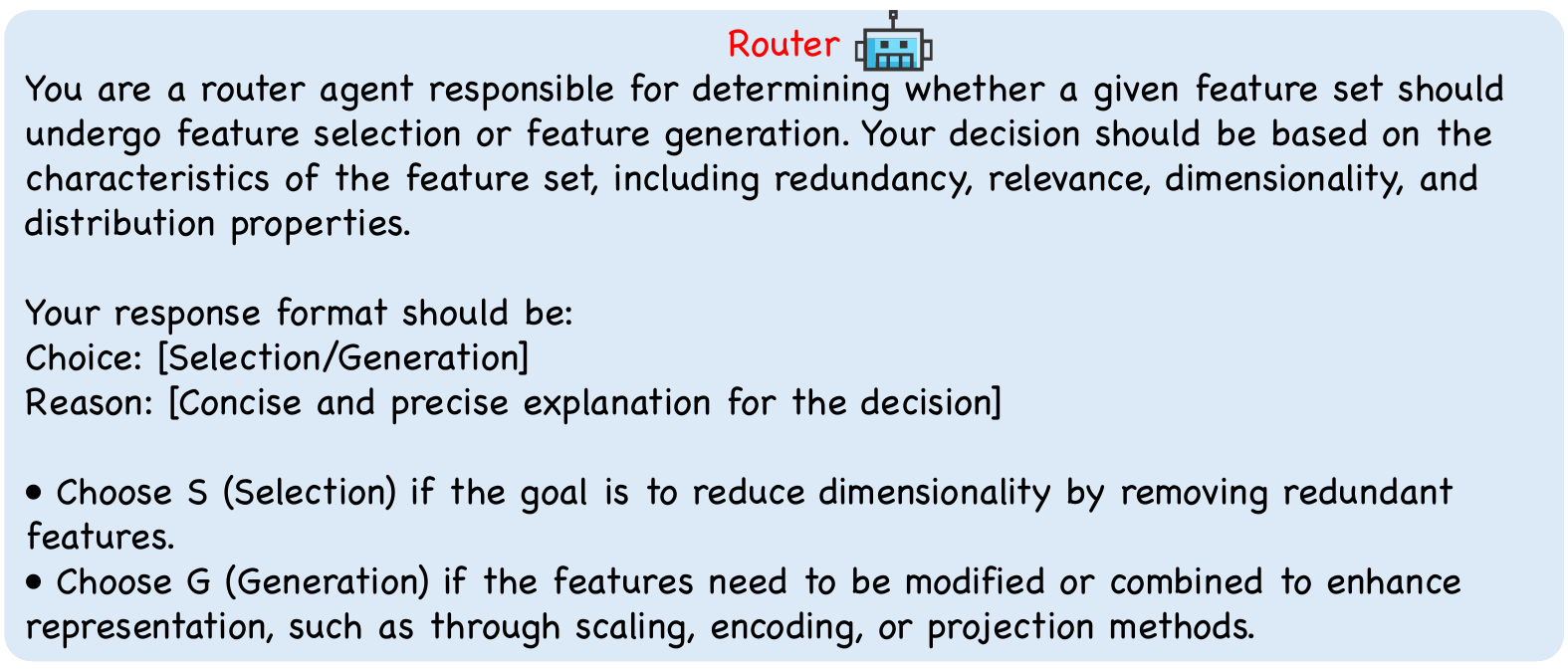}
    \caption{Problem template of router.}
    \label{fig:router}
    \vspace{-0.2cm}
\end{figure*}

\begin{figure*}[h]
    \centering
    \includegraphics[width=0.8\linewidth]{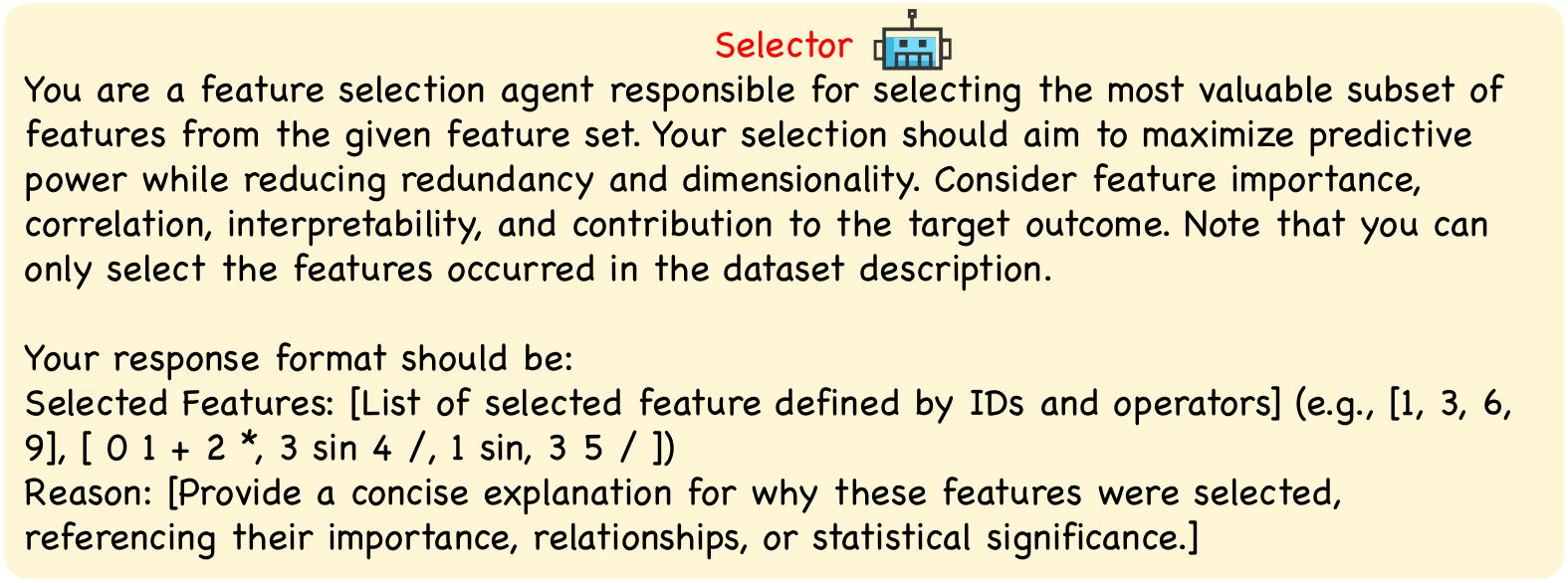}
    \caption{Problem template of selector.}
    \label{fig:selector}
    \vspace{-0.2cm}
\end{figure*}

\begin{figure*}[h]
    \centering
    \includegraphics[width=0.8\linewidth]{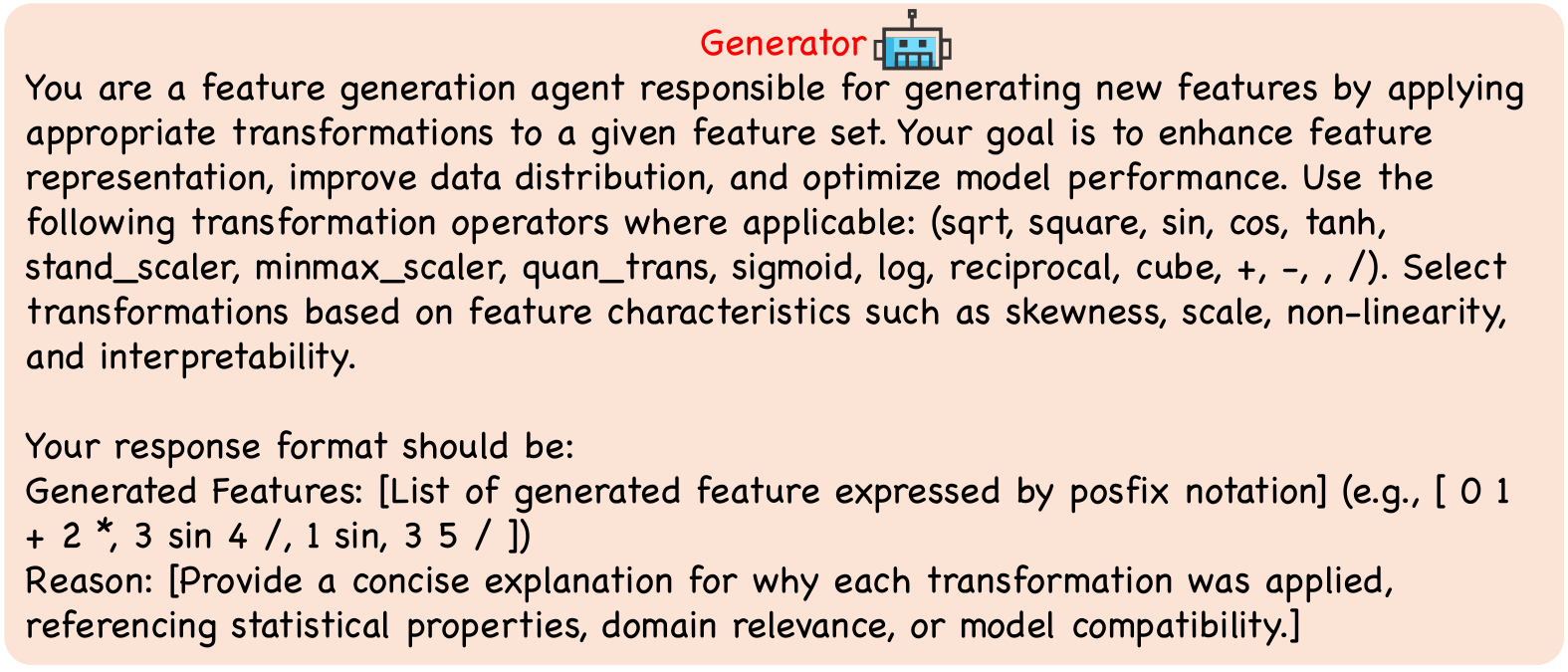}
    \caption{Problem template of generator.}
    \label{fig:generator}
    \vspace{-0.2cm}
\end{figure*}

To effectively guide the decision-making and generation behaviors of each agent, we design specialized prompt templates for the router, selector, and generator modules as shown in Figure \ref{fig:router}-\ref{fig:generator}. These templates provide structured contexts that reflect each agent’s unique functional role within the MAGS framework.

\section{Experimental Setup}
\label{ap:ex}

\subsection{Dataset Details}
We collect 6 datasets from LibSVM, UCIrvine, and OpenmlML. Table \ref{tab:dataset_stats} shows the details of the datasets.
\begin{table*}[h]
\centering
\resizebox{0.7\textwidth}{!}{%
\begin{tabular}{ccccc}
\toprule
\textbf{Dataset} & \textbf{Source} & \textbf{Task} & \textbf{Samples} & \textbf{Features} \\
\midrule
svmuide3            & LibSVM   & C & 1243 & 21    \\
german\_credit        & UCIrvine & C & 1000 & 24    \\
messidor    & UCIrvine & C & 1151 & 19    \\
\midrule
openml\_586          & OpenmlML & R & 1000 & 25    \\
openml\_589          & OpenmlML & R & 1000 & 25    \\
openml\_607          & OpenmlML & R & 1000 & 50    \\
\bottomrule
\end{tabular}
}
\caption{Dataset statistics.}
\label{tab:dataset_stats}
\vspace{-0.3cm}
\end{table*}

\subsection{Baselines}
Our baselines include 6 feature generation methods: 
1) \textbf{ERG} applies predefined operators to individual features to enrich the feature space, followed by selection.
2) \textbf{AFT} \cite{horn2020autofeat} performs iterative feature generation and combines multi-step feature selection to retain useful features.
3) \textbf{NFS} \cite{chen2019neural} models the process of generating features as sequential trajectories and optimizes them via reinforcement learning.
4) \textbf{TTG} \cite{khurana2018feature} formulates feature generation as a graph traversal problem and searches the graph using reinforcement learning.
5) \textbf{GRFG}~\cite{wang2022GRFG} generates new features via cascading agents that explore feature interactions and group-wise combinations.
6) \textbf{ELLM-FT} \cite{gong2024evolutionary} integrates evolutionary algorithm and LLM to optimize and generate new feature transformations.

We also compare the proposed method with 6 feature selection methods: 1) \textbf{KBEST} \cite{yang1997comparative} selects the top-$k$ features ranked by univariate statistical scores.
2) \textbf{LASSO} \cite{tibshirani1996regression} applies $\ell_1$ regularization to induce sparsity and eliminate irrelevant features.
3) \textbf{RFE} \cite{granitto2006recursive} recursively removes the least important features based on model weights.
4) \textbf{SARLFS} \cite{liu2021efficient} employs a single-agent reinforcement learning framework to select features by treating feature interactions as part of the state.
5) \textbf{MARLFS}~\cite{liu2019automating} assigns an agent to each feature and learns individual policies to select or deselect features based on joint rewards.
6) \textbf{FSNS} \cite{gong2025neuro} embeds feature sets into embedding space and searches for new feature sets by gradient-based method.

\subsection{Experimental Environment}
All experiments are conducted on the Ubuntu 22.04.3 LTS operating system, Intel(R) Core(TM) i9-13900KF CPU @ 3GHz, and 1 RTX 6000 Ada GPU with 48GB of RAM, using the framework of Python 3.11.4 and PyTorch 2.0.1. 

\subsection{Hyperparameters}

We collect 400 samples to fine-tune the Router and train it for 5 epochs using PPO. The search process consists of 30 iterations, each containing 6 actions, resulting in a total of 180 actions.

\section{Limitations}
While MAGS demonstrates strong performance and adaptability, several limitations remain. First, the framework incurs relatively high computational overhead due to its multi-agent structure and reliance on large language models for step-wise decision making. This may limit its applicability in latency-sensitive or resource-constrained environments.
Second, the use of in-context learning in the Transformer and Selector agents introduces scalability constraints. Since prompts must encode both feature context and demonstration examples, the method is less suitable for datasets with a large number of features, as the input may exceed the model's token limit.
Third, MAGS is inherently task-specific: it optimizes the feature space based on downstream model feedback, which enables strong performance but also means that the generated features may not generalize across tasks. In multi-task or transfer learning scenarios, additional adaptation may be required.

\end{document}